# Challenging the principle of compositionality in interpreting natural language texts

Françoise Gayral, Daniel Kayser and François Lévy
LIPN Institut Galilée
(UMR 7030 du C.N.R.S)
99 Avenue Jean-Baptiste Clément
F-93430 Villetaneuse
{fg,dk,fl}@lipn.univ-paris13.fr

This paper aims at emphasizing that, even relaxed, the hypothesis of compositionality has to face many problems when used for interpreting natural language texts. Rather than fixing these problems within the compositional framework, we believe that a more radical change is necessary, and propose another approach.

## The problems

The classical expression of compositionality, viz. the meaning of a complex expression can be obtained from the meanings of the words which compose it and from its syntactic structure, is problematic in at least four respects.

- **Enumerating senses**

As its starting point is meanings of the words, the hypothesis supposes that the meanings of all words can be explicitly enumerated.

Some formal semantics (e.g. [Mont74]) leave aside the issue of polysemy and present the compositional process as operating on words which have already been disambiguated, as if an implicit disambiguation phase had previously occurred.

Other approaches as generative systems [Pust95], while enriching the compositional process with other mechanisms (coercion, selective binding) to account for systematic polysemy, maintain nevertheless this hypothesis. However, as many works in lexical semantics have shown, it is impractical and even impossible to enumerate all the semantic values that a single word might get when the context varies. Even a simple word like "examen" considered in an academic context [GKP01] can take an arbitrary number of interpretations: event (1), process (2), physical object (paper on which the subject is written (3), paper on which the student write their answers (4)), information object (5),…

*(1) The exam will take place tomorrow.*
*(2) In this curriculum, the level of the students is controlled by a weekly exam.*
*(3) This is the operating system exam. Please make 100 copies of it.*
*(4) Paul left the exams on my desk, and asked me to grade them.*
*(5) The exam was easy.*

- **Selecting senses while getting up the syntactic tree**

Following this enumerative view of the lexicon, the interpretation process is thus reduced to a more or less combinatorial search in a finite space, paralleling syntactic constructs and semantic selection, with an order on the composition operations defined independently of the meaning.

For instance, as a verb is "higher" in the syntax tree than its arguments, the compositional hypothesis makes it possible for the verb to guide the disambiguation process of its arguments by means of selectional restrictions. But it rules out a process working the other way round, i.e. the arguments guiding the disambiguation of the verb, whereas many examples show that this frequently happens. In (6) and (7), the meaning of the verb *"couper"* (cut) depends on its subject.

(6) *L'orage a coupé la route (the storm cut the road off)* (cut = obstruct)

(7) *Ma voiture a coupé la route (my car cut the road across)* (cut = go from one side to the opposite)

in (8) and (9), on a prepositional phrase:

*(8) Suite à des inondations, les gendarmes ont coupé la route (After an important flood, the police cut the road off)* (the police cut = the police forbad the traffic)

*(9) Entraînés dans une folle poursuite, les gendarmes ont coupé la route (Involved in a pursuit, the police cut the road accross)* (the police cut = the police drove across)

The use of a purely bottom-up mechanism is also challenged by cases where several polysemic words constrain each other to finally yield a non-ambiguous interpretation. No ordering of composition of meanings can account for the equilibrium finally reached, as in these examples with *"examen"* and *"laisser"* (leave):

*(10) J'ai laissé l'examen de système sur ton bureau (I left the system exam on your desk) (leave = put, exam = paper)*

*(11) J'ai laissé l'examen de système à Paul qui est plus compétent (I left the system exam to Paul who is more competent) (leave = entrust, exam = task of writing the subject)*

*(12) J'ai laissé l'examen de système pour l'an prochain (I left the system exam for next year) (leave = postpone, exam = whole process of evaluating students)*

- **Co-presence**

Up to now, we have discussed the hypothesis of compositionality within the framework where the semantic process aims at finding a unique sense for each word of a text. But this postulate can be challenged too by cases where an occurrence of a given word takes simultaneously different meanings, a phenomenon we call co-presence, as in:

*(13) J'ai déposé l'examen de mercredi prochain sur ton bureau (I laid the exam of next Wednesday on your desk)*

In (13), *"examen"* means both an event because of its association with the date, and a physical object (paper) according to the verb.

- **Non lexical knowledge**

Last, but not least, compositionality implies that a semantic interpretation of the sentence can be built with the help of lexical and syntactic knowledge before appealing to any other factor. With this respect, it is fundamental to contrast:

*(14) La voiture passe au rouge (Lit. the car passes at red, i.e. goes through the red light)*

*(15) Le feu passe au rouge (Lit. the light passes at red, i.e. turns red).*

These sentences share the same form. Now, it is impossible to infer from the first one that the car has become red. World knowledge plays here a major role. The compositional hypothesis can either allow to take this knowledge into account, at the price of an external parameter. But if any external parameter can be accepted, the hypothesis becomes vacuous. Or it can defer the disambiguation to pragmatics. But then every issue of lexical semantics will be put sooner or later under the label of pragmatics. For instance, consider:

*la voiture dans le virage, dans le garage, dans le ravin (the car in the bend, in the garage, in the ditch).*

The words *virage, garage, ravin* belong to the same semantic category: location, whereas the inference of the position of the car relatively to this location is entirely different in each case.

Even if compositionality is restricted to grammatical semantics, i.e. if an oracle gives the meaning of each word, and the task is only to find the meaning of their assembly, it does not work much better. Consider for instance the case of plurals. Telling that a plural nominal phrase refers to a collection of individuals is not sufficient to account for collections persisting over time while their members change. Indeed, plural can be given two interpretations: a *de dicto* interpretation in which the composition of the collection varies as in (16) and a *de re* interpretation which refers to a fixed set as in (17).

*(16) For several centuries, the aborigines remained ignored*

*(17) For several hours, the aborigines remained seated to protest against their lot*

Here too, the interpretation of the grammatical feature of plural as *de dicto* or *de re* cannot be compositional, since external factors, e.g. comparison of durations, play a prominent role in this choice.

### A track for a solution

Our approach tries to avoid these questions in adopting two other hypothesis.

- **Inference**

The first one concerns the objective of interpretation: for us, it does not consist in seeking the adequate meaning of each word in a sentence but rather in getting the set of inferences that every hearer/reader considers as naturally implied by this sentence [Kay91].

- **Non-monotonic logics**

The second one concerns the means used to get these inferences. Rather than considering the interpretation process as a process by which constraints propagate one way, we consider that it is constrained both from below (e.g. the lexicon) and from above (e.g. world knowledge) so that interpretation is the result of an equilibrium reached when all the constraints are taken into account.

It is worth-noticing that the set of constraints yields not always a unique solution: even in a well-defined context, some ambiguities remain. Another feature of the problem is that most, if not all, constraints can be violated (e.g. metonymic or metaphoric readings).

Now there exists a toolbox, which deals with reaching equilibrium from a set of "soft" constraints, yielding in some cases several solutions: non-monotonic inference systems. Several of these systems have a property called semi-monotonicity; the technical definition cannot be given in this abstract; it amounts to impose that, if the "hard" knowledge stays the same, while the "soft" constraints increase, the derived consequences can only increase. As has been noticed [Brew91], this property is incompatible with putting priorities on the "soft" constraints; now it is easy to show that the phenomena described in this paper require handling such priorities. Therefore we must select a non-monotonic inference system that lacks semi-monotonicity. Reiter's semi-normal default logic ([Rei80]) is a good candidate; it allows representing the equilibrium reached by a system of constraints as a solution of a fixpoint equation.

We will show how most of the problems met by the compositional hypothesis can receive a tentative solution in the framework of a non-monotonic, non-semi-monotonic inference system.

### References

[Brew91] G. Brewka: *Cumulative Default Logic : in defense of nonmonotonic inference rules* Artificial Intelligence vol.50 n°2 pp.183-205, July 1991

[GKP01] F. Gayral, D. Kayser, N. Pernelle: *In Search of the Semantic Value(s) of an Occurrence: an example and a framework*. in Computing Meaning vol.2 (Harry Bunt, Reinhard

Muskens, Elias Thijsse, eds.) pp.53-69, vol.77 Studies in Linguistics and Philosophy, Kluwer, 2001

[Kay91] D. Kayser: *Meaning Representation vs. Knowledge Representation*. in New Inquiries into Meaning and Truth (N.Cooper et P.Engel, eds.) pp.163-186 Simon & Shuster, 1991

[Mont74] R. Montague: *Formal Philosophy: Selected Papers of Richard Montague*. Yale University Press, New Haven, 1974

[Pust95] J. Pustejovsky: *The generative lexicon*, MIT Press, Cambridge, 1995

[Rei80] R. Reiter: *A Logic for Default Reasoning*, Artificial Intelligence, vol.13 n°1-2 pp.81-132, April 1980